\documentclass[letterpaper]{article}
\usepackage[preprint]{aaai2027}
\usepackage[hyphens]{url}
\usepackage{graphicx}
\usepackage{natbib}
\usepackage{caption}
\usepackage{algorithm}
\usepackage{algorithmic}
\usepackage{booktabs}
\usepackage{amsmath}
\usepackage{amssymb}
\usepackage{placeins}
\newcommand{\method}{CoBa}
\newcommand{\strong}{CoBa-Routed-Strong}
\newcommand{\balanced}{CoBa-Routed}
\newcommand{\light}{CoBa-Routed-Light}

\title{\method{}: Cost-Effective Test-Time Scaling via Compute-Balanced Routing}
\author{
Yan Zhou\textsuperscript{\rm 1},
Yue Ouyang\textsuperscript{\rm 1},
Kaiyang Zheng\textsuperscript{\rm 1},
Suncheng Xiang\textsuperscript{\rm 2}\corresponding
}
\affiliations{
\textsuperscript{\rm 1}School of Mathematics and Statistics, Changsha University of Science and Technology, Changsha, China\\
\textsuperscript{\rm 2}School of Biomedical Engineering, Shanghai Jiao Tong University, Shanghai, China. E-mail: \texttt{xiangsuncheng17@sjtu.edu.cn}
}

\begin{document}

\maketitle

\vspace{-1em}
\begin{center}
{\footnotesize\textcolor{gray}{This paper is a preprint only; the final published version, if any, shall prevail.}}
\end{center}
\vspace{0.5em}

\begin{abstract}
Test-time scaling is often implemented by spending more compute along one axis: sampling more solutions, extending a chain of thought, or applying a stronger evaluator.  Under a fixed inference budget, these choices compete.  This paper formulates test-time reasoning as a compute-allocation problem in which a system must decide whether the next unit of compute should be spent on generation, verification, or stopping.  We introduce \method{}, a compute-balanced routing policy that first obtains a small set of candidates, applies cheap verification broadly, and routes uncertain or high-value candidates to stronger verification.  On 3,129 example--generator evaluations spanning MATH-500, AIME 2024/2025, AMC 2023, and procedural symbolic reasoning, \strong{} reaches 85.13\% macro accuracy, statistically matching a self-evaluation weighted-voting proxy at 85.20\% while using 49.1\% fewer parameter-weighted tokens.  It also matches best-of-16 majority voting within 0.01 macro-accuracy points while using 58.9\% fewer parameter-weighted tokens; paired tests retain a small best-of-16 edge at substantially higher cost.  Paired bootstrap tests show significant gains over single-sample decoding, while the remaining gap to the pool oracle exposes headroom for sharper routing.  For local reasoning systems, test-time scaling becomes a question of where the next computation is most valuable.
\end{abstract}

\section{Introduction}

Large language models can improve reasoning at test time by allocating extra computation across several actions: producing chains of thought \citep{wei2022chain}, sampling multiple solutions \citep{wang2022self}, or judging candidates with verifiers \citep{cobbe2021training,lightman2024let}.  These strategies are usually treated as separate scaling knobs: sample more, think longer, or evaluate harder.  Under a fixed inference budget, these knobs compete.  Spending the next unit of compute on one operation prevents spending it elsewhere.

\begin{figure}[!t]
  \centering
  \includegraphics[width=\columnwidth]{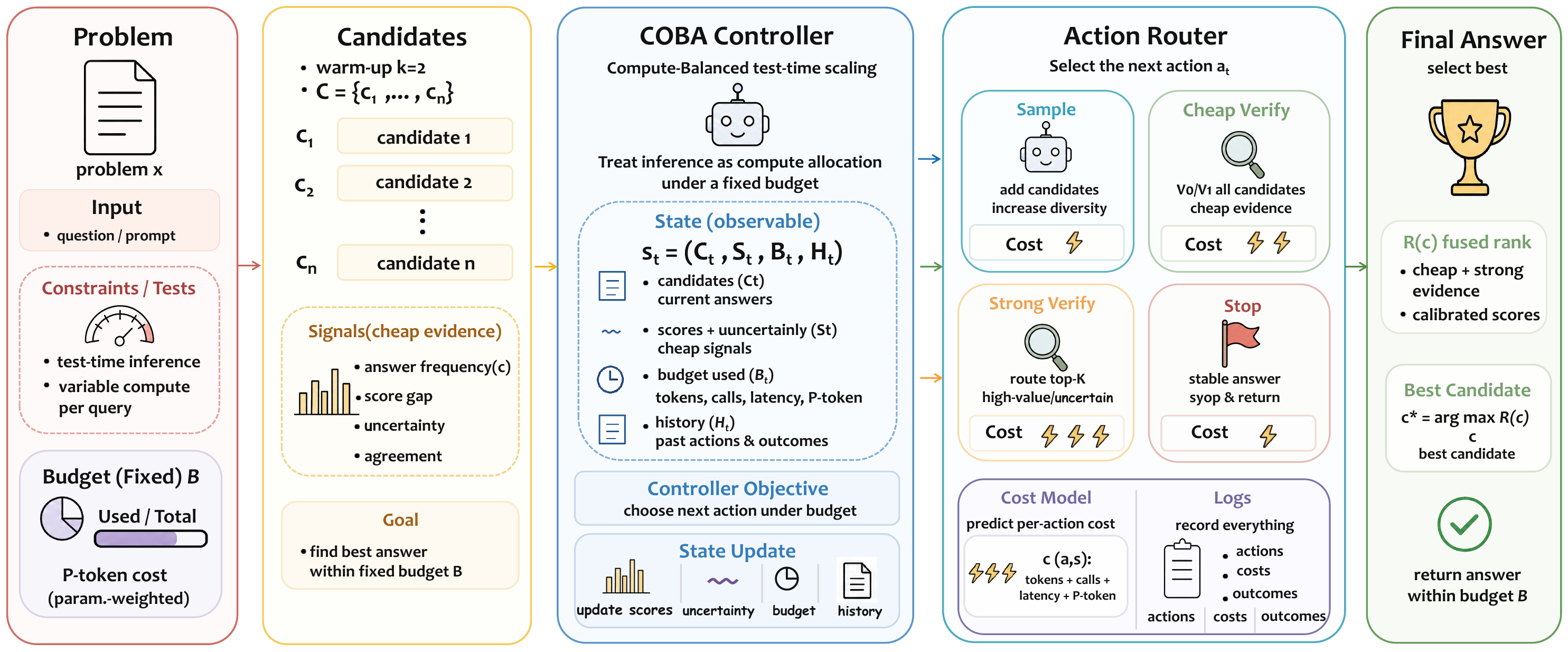}
\caption{\method{} views test-time scaling as compute allocation.  The controller routes each state among sampling, lightweight verification, strong verification, and stopping using observable uncertainty, verification, and budget features.}
  \label{fig:framework}
\end{figure}

We study test-time scaling as a compute-allocation problem: at each state, should a reasoning system sample another candidate, apply lightweight verification, invoke a stronger verifier, or stop?  \method{} (Compute-Balanced test-time scaling) answers this question with the routing framework in Figure~\ref{fig:framework}.  Its main implementation, \strong{}, obtains candidate diversity, scores candidates with cheap evidence, and routes selected candidates to stronger verification.

The central hypothesis is that accuracy gains and cost savings should be evaluated together.  A system that always samples many candidates or always invokes a strong evaluator may be accurate, yet it spends the same expensive actions on settled and ambiguous examples.  \method{} uses the same generators and verifiers available to the baselines and changes when they are used.

We evaluate \method{} with Qwen3-14B, Phi-4-reasoning, and Qwen3-8B \citep{yang2025qwen3,abdin2025phi} on competition mathematics and procedural symbolic reasoning.  \strong{} reaches the accuracy region of best-of-16 and self-evaluation weighted voting while using lower parameter-weighted cost.  Our contributions are:
\begin{itemize}
  \item We formalize test-time reasoning as a unified allocation problem over generation, verification, and stopping actions.
  \item We introduce a reproducible local routing policy with lightweight and strong verification tiers.
  \item We provide controlled replay experiments showing that adaptive routing reaches the accuracy of much more expensive sampling and evaluator-scaling baselines, with ablations and paired significance tests explaining when it helps.
\end{itemize}

\section{Related Work}

\textbf{Test-time reasoning and sampling.}
Large language models made few-shot reasoning practical at scale \citep{brown2020language,chowdhery2023palm}.  Chain-of-thought and zero-shot chain-of-thought prompting make intermediate reasoning explicit \citep{wei2022chain,kojima2022large}, while self-consistency improves robustness by sampling multiple chains and marginalizing over final answers \citep{wang2022self}.  Related test-time reasoning methods improve solutions through rationale bootstrapping, search, tool use, or self-feedback \citep{zelikman2022star,yao2022react,yao2023tree,gao2023pal,chen2022program,madaan2023self,shinn2023reflexion}.  Recent surveys organize test-time scaling by what, how, and where compute is scaled \citep{zhang2025and}; concrete methods scale inference through repeated sampling or reasoning-length control \citep{brown2024large,snell2024scaling,muennighoff2025s1}, echoing large-sample code-generation systems such as AlphaCode \citep{li2022competition}.  These methods motivate our setting and point to a finer-grained allocation question: which action should receive the next unit of compute?

\textbf{Verification and evaluation-time compute.}
Verifier-based methods train or prompt models to judge candidate solutions \citep{cobbe2021training}.  Process supervision further evaluates intermediate reasoning steps, with human-labeled and automatically constructed process signals improving mathematical verification \citep{uesato2022solving,lightman2024let,wang2024math}.  Reward-model evaluation and process-error benchmarks expose both the value and fragility of verifier signals \citep{lambert2025rewardbench,zheng2025processbench}.  Recent work also scales the evaluator itself: GenPRM increases process-reward compute through generative reasoning \citep{zhao2026genprm}, evaluation-time scaling uses reasoning models as evaluators \citep{kim2025scaling}, and unified reasoner--verifier systems couple generation with verification at inference time \citep{sareen2025putting}.  These systems make strong verification increasingly available and make verifier cost a first-order design variable.  \method{} keeps the verifier family available across comparisons and routes strong verification to selected candidates in the main local setting.

\textbf{Adaptive compute allocation.}
Adaptive allocation methods share the same motivation as ours: inputs deserve different amounts and kinds of compute.  Earlier adaptive-computation work learned how many recurrent steps to execute \citep{graves2016adaptive}; modern LLM systems route among models, cascades, or candidate ensembles to trade cost for quality \citep{chen2023frugalgpt,jiang2023llm,ong2024routellm}.  Serving- and latency-aware systems show that reasoning progress and wall-clock constraints can also shape test-time compute \citep{fu2024efficiently,huang2025latency}.  Early-stopping and monitoring frameworks such as interwhen \citep{bhat2026interwhen} further show that intermediate reasoning can be monitored before committing to a fixed endpoint.  \method{} routes among actions inside a reasoning workflow: sample another candidate, verify cheaply, verify strongly, or stop.

\textbf{Reasoning models and benchmarks.}
Specialized mathematical models and technical-reasoning systems have made competition-style evaluation increasingly demanding \citep{lewkowycz2022solving,shao2024deepseekmath,guo2025deepseek,yang2025qwen3,abdin2025phi}.  Benchmarks now cover grade-school math, competition math, graduate science, Olympiad-style reasoning, coding, and process-error detection \citep{cobbe2021training,hendrycks2021measuring,rein2023gpqa,he2024olympiadbench,jain2025livecodebench,zheng2025processbench}.  Our final experiments focus on competition mathematics and procedural symbolic reasoning, with science and code benchmarks providing natural evaluation directions for future routed systems.

\section{Problem Formulation}

Let $\mathcal{D}=\{(x_i,y_i)\}_{i=1}^{n}$ be an evaluation set, where $x_i$ is a problem and $y_i$ is used for offline evaluation.  At inference time, a system observes $x$ while $y$ remains evaluator-side.  A generator $G$ can produce candidate solutions $c=(r,a)$, where $r$ is the reasoning trace and $a$ is the extracted final answer.  A set of verifier models or rules $\mathcal{V}=\{V_0,V_1,\ldots,V_L\}$ can assign scores to candidates.  $V_0$ denotes rule-based answer frequency, $V_1$ denotes a lightweight local judge, and higher levels denote stronger but more expensive verifiers.

At decision step $t$, the system state is
\begin{equation}
s_t = (C_t, S_t, B_t, H_t),
\end{equation}
where $C_t$ is the candidate set, $S_t$ contains verifier scores observed so far, $B_t$ is the remaining budget, and $H_t$ contains history features such as answer diversity, candidate length, answer flips, and whether previous generations reached the token cap.  The action space is
\begin{equation}
\mathcal{A}=\{\mathrm{SAMPLE}, \mathrm{VERIFY}_{1}, \ldots, \mathrm{VERIFY}_{L}, \mathrm{STOP}\}.
\end{equation}
Each action has a measurable cost $c(a_t,s_t)$ in tokens, model calls, latency, and parameter-weighted tokens.

We use two cost definitions.  The total-token cost is
\begin{equation}
C_{\mathrm{tok}}=\sum_{m} \left(T^{m}_{\mathrm{in}} + T^{m}_{\mathrm{out}}\right),
\end{equation}
where $T^{m}_{\mathrm{in}}$ and $T^{m}_{\mathrm{out}}$ are input and output tokens consumed by model $m$.  The parameter-weighted token cost is
\begin{equation}
C_{\mathrm{ptok}}=\sum_{m} P_m \left(T^{m}_{\mathrm{in}}+T^{m}_{\mathrm{out}}\right),
\end{equation}
where $P_m$ is the number of model parameters in billions.  This second metric approximates the fact that a token processed by a 14B model is more expensive than a token processed by an 8B model.

For a fixed budget $B$, the allocation objective is to choose a policy $\pi(a_t\mid s_t)$ that maximizes expected task utility:
\begin{equation}
\max_{\pi} \; \mathbb{E}_{x\sim \mathcal{D}}\left[\mathbf{1}\{\hat{y}_{\pi}(x)=y\} - \lambda \frac{C_{\pi}(x)}{B}\right],
\end{equation}
where $\hat{y}_{\pi}(x)$ is the final answer selected when $\pi$ stops, $C_{\pi}(x)$ is the incurred cost, and $\lambda$ controls cost sensitivity.  The experiments instantiate this view through fixed replay policies and report the resulting accuracy--cost Pareto frontier.

\section{Method}

\subsection{\method{} Routing}

\method{} implements the policy above as a staged routing procedure, summarized in Algorithm~\ref{alg:coba}.  It first spends a small, fixed warm-up budget to obtain candidate diversity.  It then applies lightweight verification to every candidate and uses the resulting answer agreement and judge scores to decide whether to stop, sample more candidates, or route selected candidates to a strong verifier.  Ground-truth answers are never used in the policy.

The design separates two decisions that are often conflated in test-time scaling.  Candidate generation asks whether the current answer set is diverse enough to contain a plausible solution; verification asks whether the system can identify that solution from available evidence.  \method{} therefore uses cheap evidence as a triage layer before spending strong-verifier compute, making stopping and escalation explicit allocation decisions.

\begin{algorithm}[t]
\caption{\method{}-Routed Inference}
\label{alg:coba}
\begin{algorithmic}[1]
\REQUIRE problem $x$, generator $G$, verifiers $V_0,V_1,V_2,V_3$, budget $B$, warm-up count $k$, maximum candidates $N_{\max}$, strong-route count $K$
\STATE $C\leftarrow \emptyset$, $S\leftarrow \emptyset$, $b\leftarrow B$
\FOR{$j=1$ to $k$}
  \STATE Generate candidate $c_j\sim G(x)$; add $c_j$ to $C$
  \STATE Update $b$ by subtracting generation cost
\ENDFOR
\STATE Score all candidates with answer-frequency verifier $V_0$
\STATE Score all candidates with lightweight judge $V_1$
\STATE Optionally score with process verifier $V_2$ when available
\WHILE{$|C|<N_{\max}$ and $b>0$}
  \STATE Compute answer agreement, score gap, and uncertainty features
  \IF{top answer is stable and lightweight score is high}
    \STATE \textbf{break}
  \ELSE
    \STATE Generate one additional candidate and score it with $V_0,V_1$
  \ENDIF
\ENDWHILE
\STATE Rank candidates by lightweight score and answer frequency
\STATE Route the top $K$ candidates to strong verifier $V_3$
\STATE Select $\hat{c}=\arg\max_{c\in C} R(c)$ using Eq.~\eqref{eq:rank}
\RETURN extracted answer from $\hat{c}$
\end{algorithmic}
\end{algorithm}

The final ranking score combines answer frequency and verifier scores:
\begin{equation}
\label{eq:rank}
R(c)=0.20\,f(c)+0.30\,s_1(c)+0.15\,s_2(c)+0.45\,s_3(c),
\end{equation}
where $f(c)$ is the normalized frequency of the extracted answer, $s_1(c)$ is the Qwen3-8B judge score, $s_2(c)$ is the optional process-verifier score, and $s_3(c)$ is the Qwen3-14B deep-verifier score.  Sparse process scores enter only when parsed; otherwise the remaining weights are renormalized, so unrouted candidates receive neither artificial penalties nor artificial support from absent verifier evidence.

\subsection{Routing Variants}

We evaluate three variants.  \light{} uses the two-candidate warm-up and lightweight verification.  \balanced{} expands to at most four candidates and routes two candidates to the strong verifier.  \strong{} expands to at most eight candidates and routes four candidates to the strong verifier.  These variants trace a cost--accuracy curve while keeping the candidate pool and verifier definitions fixed.

Table~\ref{tab:routing-spec} gives the fixed routing specification used in the final replay.  The values were set before final aggregation with test labels reserved for evaluation.  They implement a simple allocation principle: buy a small amount of diversity first, score all visible candidates cheaply, and reserve strong verification for the few candidates most likely to change the final answer.

\begin{table}[!t]
\centering
\vspace{-0.30em}
\resizebox{\columnwidth}{!}{%
\renewcommand{\arraystretch}{1.05}
\begin{tabular}{@{}ccc@{}}
\toprule
\textbf{Component} & \textbf{Value} & \textbf{Role} \\
\midrule
Warm-up $k$ & 2 & initial diversity \\
$N_{\max}$, \light{} & 2 & no extra sampling \\
$N_{\max}$, \balanced{} & 4 & moderate routing \\
$N_{\max}$, \strong{} & 8 & high-accuracy setting \\
Strong routes $K$ & 0/2/4 & light/balanced/strong \\
Stable answer & share $\ge 0.6$, $s_1\ge 0.7$ & stop criterion \\
Cheap verifier & Qwen3-8B & all candidates \\
Strong verifier & Qwen3-14B & routed candidates \\
Fusion weights & .20/.30/.15/.45 & $f,s_1,s_2,s_3$ \\
Missing scores & renormalize & no artificial penalty \\
\bottomrule
\end{tabular}%
}
\caption{Fixed routing specification for final replay.  The policy uses answer agreement, lightweight confidence, and cached verifier features; oracle correctness is reserved for evaluation.
}
\label{tab:routing-spec}
\vspace{-0.15em}
\end{table}

\subsection{Offline Replay Protocol}

All methods operate over the same offline candidate pools.  For each dataset, generator, and token budget, we generated $N=16$ candidates per example.  Replay methods can choose prefixes, subsets, or routed verification actions from this pool; candidate correctness is reserved for the explicitly labeled oracle upper bound.  This protocol aligns best-of-$N$, evaluator scaling, difficulty-adaptive allocation, and \method{} on the same generated solutions.

Offline replay approximates methods that require interactive mid-generation continuation through labeled replay or proxy variants.  This includes the s1 budget-forcing replay, uncertainty-allocation proxy, evolving-ICL proxy, and self-evaluation weighted-voting proxy.  Direct comparisons against best-of-$N$, self-consistency, and always-on local evaluator baselines over the same pools provide the main shared-pool evidence.

\section{Experiments}

\subsection{Setup}

\textbf{Datasets.}
The main benchmark suite contains five test sets: MATH-500 \citep{hendrycks2021measuring,huggingfaceh4math500}, AIME 2024 \citep{guanning2024aime24}, AIME 2025 \citep{opencompass_aime2025}, AMC 2023 \citep{aimo2024amc}, and a hard Reasoning Gym subset \citep{stojanovski2026reasoning}.  The total evaluation contains 1,043 unique examples and 3,129 example--generator evaluations across 15 dataset--generator pairs.  For paired bootstrap tests, one dataset item with incomplete paired coverage is excluded, leaving 1,042 paired examples.

\textbf{Models.}
We use Qwen3-14B, Phi-4-reasoning, and Qwen3-8B as generators.  Qwen3-8B is also used as the lightweight yes/no judge.  Qwen3-14B is used as the strong outcome verifier.  Phi-4-reasoning provides auxiliary process-verifier scores; its parsed scores were sparse in the final run, so the main comparison centers on outcome verification.  For each dataset--generator pair, every replayed method sees the same shared pool of $N=16$ candidates generated with temperature 0.6, top-$p=0.95$, and a maximum output budget of 16,384 tokens.  The abstract budget $B$ is instantiated by the candidate, route, and token caps in Table~\ref{tab:routing-spec}; all reported costs are computed from replay metadata.  Greedy and fixed-long baselines reuse the first stored candidate from this pool, while adaptive and verifier-based methods operate on prefixes or subsets of the same stored candidates.  Under this protocol, \method{} changes which candidates are inspected and selected while drawing from the same generated evidence as the baselines.

\textbf{Baselines.}
We compare with greedy single-sample decoding, fixed-long reasoning, s1 budget-matched replay, best-of-$N$ majority voting, self-consistency, $k$-stable and overthinking-aware early stopping, difficulty-adaptive allocation, uncertainty/bandit-style allocation proxies, always-on outcome and process-plus-outcome evaluator proxies, self-evaluation weighted voting, and a pool oracle.  The pool oracle selects a correct candidate whenever one appears in the generated pool and serves as an upper bound.

\textbf{Baseline taxonomy.}
The 2025--2026 test-time scaling literature contains methods with different reproducibility states: some provide a directly replayable inference rule, while others require unavailable checkpoints, online continuation, private evaluator APIs, or task-specific training.  We separate three categories.  \emph{Direct baselines} such as best-of-$N$ and self-consistency are implemented literally over the shared pool.  \emph{Local proxies} preserve the inference-time pattern of recent methods, such as always-on reasoning evaluation, unified reasoner--verifier weighted voting, and monitor-style early stopping, and are labeled as proxies when exact public artifacts are unavailable.  \emph{Upper bounds} such as the pool oracle use evaluator-side information.  This taxonomy anchors the main comparison: \method{} and its baselines operate under the same local candidate and verifier evidence, so frontier movement reflects allocation under shared model access.

\textbf{Metrics.}
We report Accuracy~$\uparrow$, average total tokens~$\downarrow$, average model calls~$\downarrow$, average parameter-weighted tokens~$\downarrow$, and average measured latency in seconds~$\downarrow$.  Arrows indicate whether higher or lower values are better.  Accuracy is computed after enhanced answer extraction and task-specific normalization.  Mathematical answers are normalized for boxes, fractions, tuples, $\pi$, and equivalent symbolic forms where supported; raw generated text is preserved for audit.  Answer extraction is important because routing changes which candidates survive to final selection.  We apply the same enhanced extractor to every method after generation, prioritizing boxed answers, explicit final-answer statements, and final symbolic expressions, while logging re-extracted answers, extraction status, token-cap hits, and malformed boxes for audit.  This keeps the comparison centered on allocation under shared post-processing.

\textbf{Hardware and serving.}
Experiments were run locally on RTX 3090 GPUs using FP16 or quantized serving where necessary.  We use vLLM for OpenAI-compatible local serving \citep{kwon2023efficient}; the reported model calls, tokens, parameter-weighted tokens, and measured latency come from replay metadata.

\textbf{Statistics.}
We use paired bootstrap tests over examples with 2,000 resamples.  A comparison is marked significant when the 95\% confidence interval of the paired accuracy difference excludes zero.  Because all methods are replayed on the same candidate pools, paired tests compare decisions under identical generated evidence.

\subsection{Main Accuracy--Cost Results}

Table~\ref{tab:main} reports macro-averaged results across the 15 dataset--generator pairs.  \strong{} obtains 85.13\% accuracy with $5.80\times 10^4$ total tokens and $6.30\times 10^5$ parameter-weighted tokens per item.  This is statistically comparable to the self-evaluation weighted-voting proxy, which obtains 85.20\%, but \strong{} uses 49.1\% fewer parameter-weighted tokens.  \strong{} also matches best-of-16 majority voting within 0.01 macro-accuracy points while reducing parameter-weighted tokens by 58.9\%.

\begin{table*}[!t]
\centering
\small
\setlength{\tabcolsep}{4.6pt}
\renewcommand{\arraystretch}{1.05}
\begin{tabular*}{\textwidth}{@{\extracolsep{\fill}}cccccc@{}}
\toprule
\multicolumn{1}{c}{} & \multicolumn{1}{c}{\textbf{Quality}} & \multicolumn{4}{c}{\textbf{Cost / Efficiency}} \\
\cmidrule(lr){2-2}\cmidrule(lr){3-6}
\textbf{Method} & \textbf{Acc.~$\uparrow$} & \textbf{Total Tok.~$\downarrow$} & \textbf{Calls~$\downarrow$} & \textbf{P-Tok.~$\downarrow$} & \textbf{Latency~$\downarrow$} \\
\midrule
Greedy / fixed long & 75.72 & 8.0k & 1.00 & 94.4k & 200.8 \\
\light{} & 78.88 & 24.1k & 3.97 & 252.9k & 409.1 \\
Difficulty-adaptive proxy & 80.01 & 64.3k & 11.47 & 717.8k & 1005.2 \\
Best-of-4 majority & 80.63 & 32.6k & 4.00 & 382.8k & 817.7 \\
\balanced{} & 82.92 & 36.4k & 5.48 & 392.4k & 612.2 \\
$k$-stable stop & 82.93 & 39.4k & 3.54 & 460.2k & 1018.0 \\
Overthinking stop & 83.50 & 28.2k & 2.89 & 333.1k & 721.8 \\
Best-of-8 majority & 84.53 & 65.3k & 8.00 & 767.6k & 1638.7 \\
\strong{} & 85.13 & 58.0k & 7.91 & 629.9k & 1013.3 \\
\midrule
Self-eval weighted proxy & \textbf{85.20} & 112.6k & 19.87 & 1236.7k & 1670.5 \\
Best-of-16 majority & 85.12 & 130.5k & 16.00 & 1533.3k & 3269.7 \\
\midrule
Pool oracle & 91.36 & 15.6k & 1.53 & 181.0k & 398.3 \\
\bottomrule
\end{tabular*}
\caption{Main macro results over 3,129 example--generator evaluations aggregated across 15 dataset--generator pairs.  Param-Tok denotes billion-parameter-weighted tokens.  The oracle is an evaluator-side diagnostic upper bound.}
\label{tab:main}
\end{table*}

\begin{figure}[!t]
  \centering
  \includegraphics[width=\columnwidth]{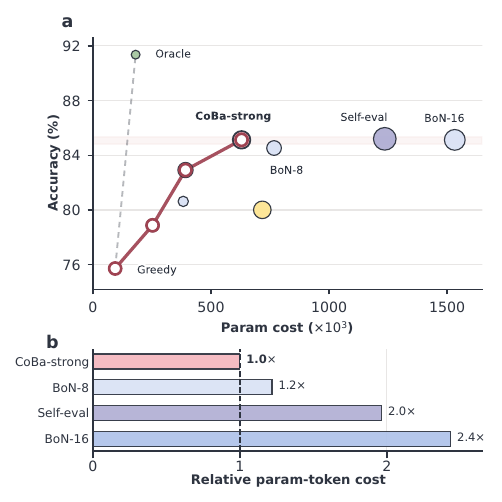}
  \caption{Accuracy--cost frontier.  \method{} reaches the high-accuracy region with lower cost.}
  \label{fig:pareto}
\end{figure}

Figure~\ref{fig:pareto} shows the same result as a frontier view.  The \method{} variants occupy the middle-to-upper part of the frontier: \light{} is inexpensive, \balanced{} improves over best-of-4 at similar cost, and \strong{} reaches the accuracy of much more expensive sampling and self-evaluation baselines.

Marginal accuracy differences in Table~\ref{tab:main} are macro-averaged over dataset--generator pairs, whereas the paired-bootstrap comparisons reported later are per-example.  The two can therefore differ when methods behave similarly on large datasets (e.g., MATH-500) and diverge on smaller ones.

\subsection{Results by Dataset}

Table~\ref{tab:dataset} shows that routed verification helps most on hard contest settings, where the system benefits from additional candidates and strong verification, while still maintaining high performance on MATH-500 and Reasoning Gym.  On AIME 2024, \strong{} improves from 65.6\% greedy accuracy to 82.2\%.  On AIME 2025, the large oracle gap points to candidate generation as the dominant remaining source of improvement.

\begin{table}[!t]
\centering
\vspace{-0.35em}
\resizebox{\columnwidth}{!}{%
\renewcommand{\arraystretch}{1.05}
\begin{tabular}{@{}cccccc@{}}
\toprule
\multicolumn{1}{c}{} & \multicolumn{1}{c}{\textbf{General}} & \multicolumn{3}{c}{\textbf{Contest}} & \multicolumn{1}{c}{\textbf{Procedural}} \\
\cmidrule(lr){2-2}\cmidrule(lr){3-5}\cmidrule(lr){6-6}
\textbf{Method} & \textbf{MATH~$\uparrow$} & \textbf{A24~$\uparrow$} & \textbf{A25~$\uparrow$} & \textbf{AMC~$\uparrow$} & \textbf{RGym~$\uparrow$} \\
\midrule
Greedy & 86.4 & 65.6 & 53.3 & 85.1 & 88.2 \\
Best-of-8 & 87.9 & 77.8 & \textbf{73.3} & 91.2 & 92.5 \\
Best-of-16 & \textbf{88.3} & 80.0 & 72.2 & 91.6 & \textbf{93.5} \\
Self-eval weighted & 87.8 & 81.1 & 72.2 & \textbf{92.4} & 92.5 \\
\strong{} & 87.6 & \textbf{82.2} & 71.1 & \textbf{92.4} & 92.3 \\
\midrule
Pool oracle & 90.3 & 87.8 & 83.3 & 95.6 & 99.8 \\
\bottomrule
\end{tabular}%
}
\caption{Per-dataset accuracy averaged over the three generators.  Bold marks the best non-oracle method; the oracle row is an evaluator-side upper bound.  A24 and A25 denote AIME 2024 and AIME 2025; RGym denotes Reasoning Gym hard.}
\label{tab:dataset}
\vspace{-0.15em}
\end{table}

The per-dataset pattern separates allocation gains from candidate-pool headroom.  \strong{} consistently improves over greedy decoding and is competitive with moderate-cost sampling or self-evaluation; the oracle remains substantially higher on AIME 2025 and Reasoning Gym.  Routing therefore improves the use of available compute, while the hardest examples highlight the value of better candidate generation and stronger uncertainty estimates.  A compact heatmap of the same table is provided in the supplementary material.

The deployment implication is that the savings are distributed across the benchmark.  Easier examples often stop after cheap agreement and lightweight judging, whereas harder contest problems consume additional samples and strong-verifier calls.  Uniform best-of-$N$ or always-on evaluation pays the hard-example budget on every example; routing concentrates that budget where the decision can still change.

The contrast between AIME 2024 and AIME 2025 is especially informative.  On AIME 2024, additional candidate diversity and routed verification are often enough to recover from an initially weak sample.  On AIME 2025, the oracle gap remains large.  Some examples contain a correct answer in the pool but lack verification signals strong enough to surface it; others lack a correct candidate in the pool.  This separates two failure modes hidden by a single aggregate score: allocation errors, where better routing could select an existing correct candidate, and generation errors, where the next useful unit of compute should create better candidates before scoring the current pool more deeply.  When the candidate pool already contains a correct answer, routing can improve the final decision by concentrating strong verification on the few candidates whose answers still compete.  When the pool itself is wrong, deeper evaluator passes mostly audit bad candidates more carefully.  The large AIME 2025 oracle gap therefore maps the next improvement regime: better candidate creation must be paired with selective scoring of the current pool.

\subsection{Evidence for Adaptive Allocation}

\begin{figure}[!t]
  \centering
  \includegraphics[width=0.94\columnwidth]{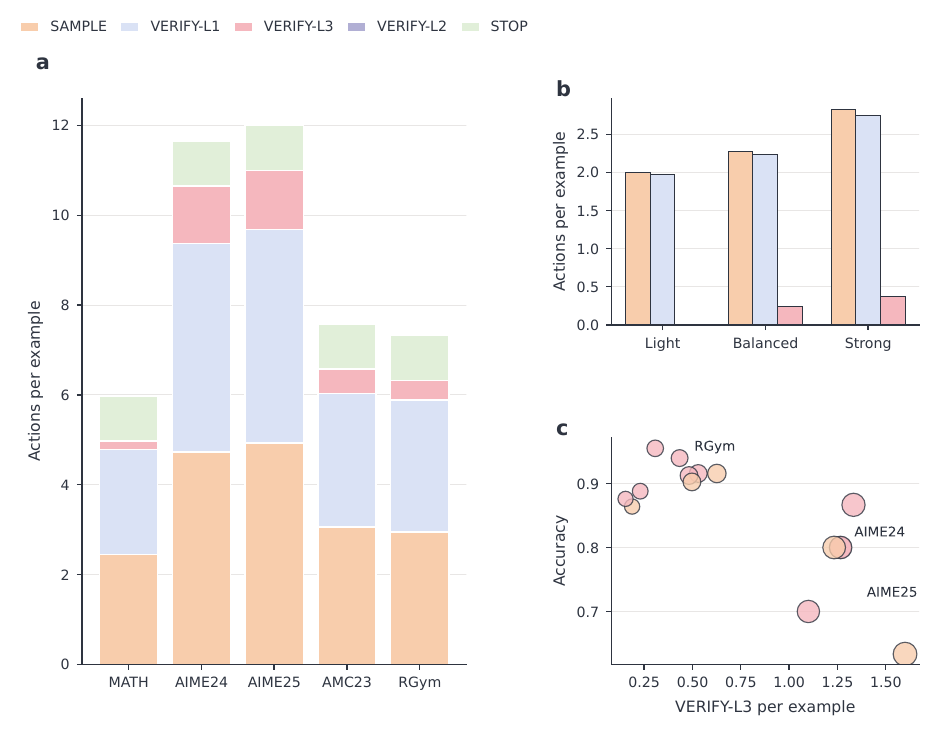}
  \caption{Adaptive action allocation.  Harder datasets receive more sampling and strong verification, while easier or more stable datasets stop earlier after lightweight evidence.}
  \label{fig:actions}
\end{figure}

The central argument is that different examples should receive different actions.  Figure~\ref{fig:actions} shows this directly: routing uses stronger verification and additional sampling unevenly across datasets, producing an action mix that differs from a uniform compute cap.  The action mix provides the behavioral evidence behind the frontier result.

The same pattern appears in the per-dataset frontiers retained in the supplementary material: routed policies move toward the high-accuracy, lower-cost region on MATH-500, AMC, AIME, and Reasoning Gym, with AIME 2025 exposing the largest remaining gap.  The routing policy changes the composition of compute across datasets: easy or stable examples terminate after agreement and lightweight judging, moderately ambiguous examples receive a few extra candidates, and the most decision-relevant candidates are escalated to strong verification.  That action distribution is the behavioral evidence behind the frontier improvement: expensive actions are reallocated away from already-settled cases and toward decisions that can still change.

Verifier scaling and generator scaling expose complementary bottlenecks.  A sampler can discover a correct answer and still leave the answer distribution split.  An always-on evaluator improves selection and spends the same effort on settled and ambiguous candidates.  Routing provides the missing middle layer: cheap evidence estimates whether the decision is settled, and strong verification is reserved for candidates where the final answer can still change.

\subsection{Significance and Ablations}

Paired bootstrap tests support three conclusions.  First, \strong{} significantly improves over greedy decoding: the paired accuracy difference is $+3.74$ points with a 95\% confidence interval of $[+2.97,+4.54]$ points.  Second, \strong{} is statistically indistinguishable from the self-evaluation weighted-voting proxy: the observed difference is $-0.16$ points and the confidence interval includes zero.  Third, best-of-16 remains slightly more accurate than \strong{} under the paired test, but the difference is only $0.70$ points and comes at 2.43$\times$ the parameter-weighted cost.

Two controls keep the comparison centered on allocation.  First, enhanced extraction is applied uniformly to every method after generation, and raw model outputs are kept unchanged.  Second, oracle labels are reserved for final evaluation and the explicitly marked pool-oracle upper bound.  The routing features available to \method{} are answer agreement, local verifier scores, candidate lengths, token-cap indicators, repetition signals, and remaining budget.

\begin{table}[!t]
\centering
\vspace{-0.35em}
\resizebox{\columnwidth}{!}{%
\renewcommand{\arraystretch}{1.05}
\begin{tabular}{@{}ccccc@{}}
\toprule
\multicolumn{1}{c}{} & \multicolumn{1}{c}{\textbf{Quality}} & \multicolumn{2}{c}{\textbf{Cost}} & \multicolumn{1}{c}{\textbf{Policy}} \\
\cmidrule(lr){2-2}\cmidrule(lr){3-4}\cmidrule(lr){5-5}
\textbf{Variant} & \textbf{Acc.~$\uparrow$} & \textbf{Calls~$\downarrow$} & \textbf{P-Tok~$\downarrow$} & \textbf{Setting} \\
\midrule
\light{} & 78.88 & 3.97 & 252.9k & L1 only \\
\balanced{} & 82.92 & 5.48 & 392.4k & top-2 \\
\strong{} & 85.13 & 7.91 & 629.9k & top-4 \\
\midrule
Self-eval & \textbf{85.20} & 19.87 & 1236.7k & all \\
\midrule
Oracle & 91.36 & 1.53 & 181.0k & upper \\
\bottomrule
\end{tabular}%
}
\caption{Routing-strength ablation.  P-Tok denotes parameter-weighted tokens.}
\label{tab:ablation}
\vspace{-0.15em}
\end{table}

Table~\ref{tab:ablation} isolates routing strength.  Moving from \light{} to \balanced{} adds a small number of strong-verifier calls and yields a large accuracy gain, indicating that the first routed candidates are high value.  Moving from \balanced{} to \strong{} continues to improve accuracy at a higher marginal cost.  This diminishing-return pattern is exactly what a compute-allocation view predicts: verification is useful, and its value depends on which candidate receives it.  The self-evaluation baseline reaches the same accuracy region through broad evaluation, while \strong{} obtains nearly the same endpoint through selective evaluation.

The learned MLP controller was also evaluated and degenerated to a near-greedy policy in the final leave-one-dataset-out setting.  This negative result is useful: the current evidence supports transparent routed allocation, while learning a robust controller from the available offline trajectories remains open.

Paired bootstrap tests clarify the accuracy--cost tradeoff.  \strong{} significantly improves over greedy decoding by $+3.74$ points (95\% CI $[+2.97,+4.54]$, $p<.001$), is statistically indistinguishable from best-of-8 and the self-evaluation proxy, and remains slightly below best-of-16 by $0.70$ points (95\% CI $[-1.25,-0.16]$, $p=.004$).  The key tradeoff is the price of the final accuracy margin: best-of-16 uses 2.43$\times$ more parameter-weighted tokens.  We read this as a Pareto improvement over common moderate-cost baselines and as a cost reduction at the high-accuracy end.

\begin{table}[!t]
\centering
\vspace{-0.35em}
\resizebox{0.94\columnwidth}{!}{%
\renewcommand{\arraystretch}{1.05}
\begin{tabular}{@{}ccccc@{}}
\toprule
\multicolumn{1}{c}{} & \multicolumn{1}{c}{\textbf{Quality}} & \multicolumn{3}{c}{\textbf{Cost}} \\
\cmidrule(lr){2-2}\cmidrule(lr){3-5}
\textbf{Method} & \textbf{Acc.~$\uparrow$} & \textbf{Calls~$\downarrow$} & \textbf{Tok.~$\downarrow$} & \textbf{Rel.~cost~$\downarrow$} \\
\midrule
\strong{} & 85.13 & 7.91 & 58.0k & 1.00$\times$ \\
Self-eval & \textbf{85.20} & 19.87 & 112.6k & 1.96$\times$ \\
Best-of-16 & 85.12 & 16.00 & 130.5k & 2.43$\times$ \\
\bottomrule
\end{tabular}%
}
\caption{Cost among high-accuracy methods.  Relative cost is computed against \strong{} using parameter-weighted tokens.}
\label{tab:highcost}
\vspace{-0.15em}
\end{table}

\subsection{Where the Cost Is Saved}

Table~\ref{tab:highcost} isolates the high-accuracy regime, where the main distinction is the amount of extra compute needed for the final gain.  Best-of-16 spends on diversity by generating sixteen full candidates, even when answers already agree.  Self-evaluation weighted voting spends on evaluation by judging many candidates, including candidates whose answers can be decided from cheap agreement.  \strong{} sits between these extremes: it first buys modest diversity, then uses cheap evidence to decide which candidates deserve stronger verification.

The savings therefore come from two places.  First, \method{} avoids repeating expensive actions on settled examples: when sampled answers agree and the lightweight judge is confident, routing stops early before a fixed evaluator stack would be invoked.  Second, it concentrates strong verification on the few candidates most likely to change the final decision.  This is why \strong{} approaches self-evaluation accuracy while using fewer calls and roughly half the token cost.  The cost reduction remains meaningful even against baselines that are already adaptive in a loose sense.  Best-of-$N$ adapts through diversity, because every extra sample is another full candidate.  Broad self-evaluation adapts after generating many candidates, because nearly all of them are judged.  \method{} adapts along both axes: it buys a limited amount of diversity first, then uses cheap evidence to decide whether more generation or stronger verification is the better next action.

The remaining gap is also informative.  On AIME 2025 and some symbolic tasks, the oracle often finds a correct candidate that routing fails to select.  These cases require sharper uncertainty or process signals, since extra always-on verifier calls mainly rescore the same ambiguous pool.

\section{Discussion}

The results support the compute-allocation view of test-time scaling.  Best-of-$N$ and self-consistency improve accuracy by buying diversity and spending uniformly across examples.  Strong evaluator baselines improve selection and pay evaluator cost for many candidates that can be decided cheaply.  \method{} benefits from the same generated candidates and verifier models, and obtains a better cost profile by routing strong compute after cheap evidence suggests it is useful.

The practical implication is that routing helps most when the candidate pool is informative and unevenly difficult.  Many examples can be settled after a few generations plus lightweight judging, so broad sampling or strong evaluation spends budget on cases that are already decided.  The AIME 2025 results mark the complementary regime: when the pool often lacks a correct answer, the next gain comes from stronger candidate creation together with better allocation.

The strongest evidence is the combination of frontier movement, action-distribution shifts, and cost-controlled significance tests.  Best-of-16 remains slightly ahead in paired accuracy, and \strong{} reaches the same practical accuracy region with much lower compute.  This is the regime most relevant to local inference deployments, where moderate sampling is feasible and deep evaluation of every candidate is costly.

This also suggests a reporting norm for test-time scaling systems.  Accuracy alone can hide whether a method buys better candidates, spends more verifier effort, or simply stops earlier.  Reporting the action mix, the cost--accuracy frontier, and the oracle gap together makes the allocation mechanism falsifiable: a claimed routing gain should show both where compute was removed and where useful compute was concentrated.

The same evidence also indicates how the allocation view should expand next.  The present benchmark suite concentrates on mathematical and procedural symbolic reasoning; natural science and code generation bring longer horizons, heterogeneous tools, and richer public checks.  The proxy baselines retain the inference-time patterns of recent adaptive systems under local replay, allowing shared-pool comparisons while exposing where public checkpoints, online continuation, or service evaluators would introduce new system variables.  Measured latency already enters the audit, and future controllers can make it a routed objective alongside accuracy and token cost.  Learned controllers, which currently trail the transparent policy, offer another route as process signals become denser.  In the present shared-pool setting, selective allocation moves the accuracy--cost frontier favorably, while stronger generators, richer process signals, and online continuation define the next systems layer.

\section{Conclusion}

This paper reframes test-time scaling as compute allocation.  \method{} asks which action deserves the next unit of compute across generation, verification, and stopping.  Across three local generators and five reasoning benchmarks, routed generation--verification allocation matches much more expensive sampling and self-evaluation baselines while using substantially fewer parameter-weighted tokens.  The remaining oracle gap turns allocation into a productive axis for practical reasoning systems: better candidate creation, sharper uncertainty, and richer process signals can all be expressed as decisions about the next computation.  More broadly, the results suggest that future test-time systems should treat generation and verification as a shared budget to be allocated.  A practical scaling system should expose its final answer together with the computation path used to reach it.

The allocation trace also makes failures actionable.  When a correct candidate appears in the pool but routing misses it, the next computation should improve uncertainty estimation or verification.  When the pool lacks a correct candidate, the next computation should expand generation.  This distinction turns oracle gaps from retrospective scores into design signals for future reasoning systems.

Taken together, the traces point to operational test-time scaling: a reasoner should expose the answer, the computation path, and the marginal action that changed the decision, connecting model capability, verifier evidence, latency, and user value.

\FloatBarrier
\newpage
\bibliography{paper}

@article{wei2022chain,
  title = {Chain-of-thought prompting elicits reasoning in large language models},
  author = {Wei, Jason and Wang, Xuezhi and Schuurmans, Dale and Bosma, Maarten and Xia, Fei and Chi, Ed and Le, Quoc V and Zhou, Denny and others},
  journal = {Advances in neural information processing systems},
  volume = {35},
  pages = {24824--24837},
  year = {2022}
}

@article{wang2022self,
  title = {Self-consistency improves chain of thought reasoning in language models},
  author = {Wang, Xuezhi and Wei, Jason and Schuurmans, Dale and Le, Quoc and Chi, Ed and Narang, Sharan and Chowdhery, Aakanksha and Zhou, Denny},
  journal = {arXiv preprint arXiv:2203.11171},
  year = {2022}
}

@article{cobbe2021training,
  title = {Training verifiers to solve math word problems, 2021},
  author = {Cobbe, Karl and Kosaraju, Vineet and Bavarian, Mohammad and Chen, Mark and Jun, Heewoo and Kaiser, Lukasz and Plappert, Matthias and Tworek, Jerry and Hilton, Jacob and Nakano, Reiichiro and others},
  journal = {URL https://arxiv. org/abs/2110.14168},
  volume = {9},
  year = {2021}
}

@inproceedings{lightman2024let,
  title = {Let's verify step by step},
  author = {Lightman, Hunter and Kosaraju, Vineet and Burda, Yuri and Edwards, Harrison and Baker, Bowen and Lee, Teddy and Leike, Jan and Schulman, John and Sutskever, Ilya and Cobbe, Karl},
  booktitle = {International Conference on Learning Representations},
  volume = {2024},
  pages = {39578--39601},
  year = {2024}
}

@article{snell2024scaling,
  title = {Scaling llm test-time compute optimally can be more effective than scaling model parameters},
  author = {Snell, Charlie and Lee, Jaehoon and Xu, Kelvin and Kumar, Aviral},
  journal = {arXiv preprint arXiv:2408.03314},
  year = {2024}
}

@inproceedings{muennighoff2025s1,
  title = {s1: Simple test-time scaling},
  author = {Muennighoff, Niklas and Yang, Zitong and Shi, Weijia and Li, Xiang Lisa and Fei-Fei, Li and Hajishirzi, Hannaneh and Zettlemoyer, Luke and Liang, Percy and Cand{\`e}s, Emmanuel and Hashimoto, Tatsunori B},
  booktitle = {Proceedings of the 2025 Conference on Empirical Methods in Natural Language Processing},
  pages = {20286--20332},
  year = {2025}
}

@article{yang2025qwen3,
  title = {Qwen3 technical report},
  author = {Yang, An and Li, Anfeng and Yang, Baosong and Zhang, Beichen and Hui, Binyuan and Zheng, Bo and Yu, Bowen and Gao, Chang and Huang, Chengen and Lv, Chenxu and others},
  journal = {arXiv preprint arXiv:2505.09388},
  year = {2025}
}

@inproceedings{zhao2026genprm,
  title = {Genprm: Scaling test-time compute of process reward models via generative reasoning},
  author = {Zhao, Jian and Liu, Runze and Zhang, Kaiyan and Zhou, Zhimu and Gao, Junqi and Li, Dong and Lyu, Jiafei and Qian, Zhouyi and Qi, Biqing and Li, Xiu and others},
  booktitle = {Proceedings of the AAAI Conference on Artificial Intelligence},
  volume = {40},
  pages = {34932--34940},
  year = {2026}
}

@article{kim2025scaling,
  title = {Scaling evaluation-time compute with reasoning models as process evaluators},
  author = {Kim, Seungone and Wu, Ian and Lee, Jinu and Yue, Xiang and Lee, Seongyun and Moon, Mingyeong and Gashteovski, Kiril and Lawrence, Carolin and Hockenmaier, Julia and Neubig, Graham and others},
  journal = {arXiv preprint arXiv:2503.19877},
  year = {2025}
}

@article{sareen2025putting,
  title = {Putting the value back in rl: Better test-time scaling by unifying llm reasoners with verifiers},
  author = {Sareen, Kusha and Moss, Morgane M and Sordoni, Alessandro and Agarwal, Rishabh and Hosseini, Arian},
  journal = {arXiv preprint arXiv:2505.04842},
  year = {2025}
}

@article{bhat2026interwhen,
  title = {interwhen: A Generalizable Framework for Steering Reasoning Models with Test-time Verification},
  author = {Bhat, Vishak K and Chanda, Prateek and Ekbote, Vijval and Khandelwal, Ashmit and Swaroop, Maitreyi and Balasubramanian, Vineeth N and Kambhampati, Subbarao and Natarajan, Nagarajan and Sharma, Amit},
  journal = {arXiv preprint arXiv:2602.11202},
  year = {2026}
}

@article{abdin2025phi,
  title = {Phi-4-reasoning technical report},
  author = {Abdin, Marah and Agarwal, Sahaj and Awadallah, Ahmed and Balachandran, Vidhisha and Behl, Harkirat and Chen, Lingjiao and de Rosa, Gustavo and Gunasekar, Suriya and Javaheripi, Mojan and Joshi, Neel and others},
  journal = {arXiv preprint arXiv:2504.21318},
  year = {2025}
}

@article{hendrycks2021measuring,
  title = {Measuring mathematical problem solving with the math dataset},
  author = {Hendrycks, Dan and Burns, Collin and Kadavath, Saurav and Arora, Akul and Basart, Steven and Tang, Eric and Song, Dawn and Steinhardt, Jacob},
  journal = {arXiv preprint arXiv:2103.03874},
  year = {2021}
}

@misc{huggingfaceh4math500,
  title = {{MATH-500}},
  author = {{Hugging Face H4}},
  year = {2023},
  howpublished = {\url{https://huggingface.co/datasets/HuggingFaceH4/MATH-500}},
  note = {Accessed 2026-06-20}
}

@misc{guanning2024aime24,
  title = {{AIME} 2024},
  author = {Guanning},
  year = {2024},
  howpublished = {\url{https://huggingface.co/datasets/guanning/aime2024}},
  note = {Accessed 2026-06-20}
}

@misc{opencompass_aime2025,
  title = {AIME 2025 Dataset},
  author = {{OpenCompass}},
  year = {2025},
  note = {Hugging Face Dataset Repository},
  url = {https://huggingface.co/datasets/opencompass/AIME2025}
}

@misc{aimo2024amc,
  title = {{AIMO} Validation {AMC}},
  author = {{AI-MO}},
  year = {2024},
  howpublished = {\url{https://huggingface.co/datasets/AI-MO/aimo-validation-amc}},
  note = {Accessed 2026-06-20}
}

@article{stojanovski2026reasoning,
  title = {Reasoning gym: Reasoning environments for reinforcement learning with verifiable rewards},
  author = {Stojanovski, Zafir and Stanley, Oliver and Sharratt, Joe and Jones, Richard and Adefioye, Abdulhakeem and Kaddour, Jean and K{\"o}pf, Andreas},
  journal = {Advances in Neural Information Processing Systems},
  volume = {38},
  year = {2026}
}

@inproceedings{kwon2023efficient,
  title = {Efficient memory management for large language model serving with pagedattention},
  author = {Kwon, Woosuk and Li, Zhuohan and Zhuang, Siyuan and Sheng, Ying and Zheng, Lianmin and Yu, Cody Hao and Gonzalez, Joseph and Zhang, Hao and Stoica, Ion},
  booktitle = {Proceedings of the 29th symposium on operating systems principles},
  pages = {611--626},
  year = {2023}
}

@article{brown2020language,
  title = {Language models are few-shot learners},
  author = {Brown, Tom and Mann, Benjamin and Ryder, Nick and Subbiah, Melanie and Kaplan, Jared D and Dhariwal, Prafulla and Neelakantan, Arvind and Shyam, Pranav and Sastry, Girish and Askell, Amanda and others},
  journal = {Advances in neural information processing systems},
  volume = {33},
  pages = {1877--1901},
  year = {2020}
}

@article{chowdhery2023palm,
  title = {Palm: Scaling language modeling with pathways},
  author = {Chowdhery, Aakanksha and Narang, Sharan and Devlin, Jacob and Bosma, Maarten and Mishra, Gaurav and Roberts, Adam and Barham, Paul and Chung, Hyung Won and Sutton, Charles and Gehrmann, Sebastian and others},
  journal = {Journal of machine learning research},
  volume = {24},
  number = {240},
  pages = {1--113},
  year = {2023}
}

@article{kojima2022large,
  title = {Large language models are zero-shot reasoners},
  author = {Kojima, Takeshi and Gu, Shixiang Shane and Reid, Machel and Matsuo, Yutaka and Iwasawa, Yusuke},
  journal = {Advances in neural information processing systems},
  volume = {35},
  pages = {22199--22213},
  year = {2022}
}

@article{zelikman2022star,
  title = {Star: Bootstrapping reasoning with reasoning},
  author = {Zelikman, Eric and Wu, Yuhuai and Mu, Jesse and Goodman, Noah},
  journal = {Advances in Neural Information Processing Systems},
  volume = {35},
  pages = {15476--15488},
  year = {2022}
}

@article{yao2022react,
  title = {React: Synergizing reasoning and acting in language models},
  author = {Yao, Shunyu and Zhao, Jeffrey and Yu, Dian and Du, Nan and Shafran, Izhak and Narasimhan, Karthik and Cao, Yuan},
  journal = {arXiv preprint arXiv:2210.03629},
  year = {2022}
}

@article{yao2023tree,
  title = {Tree of thoughts: Deliberate problem solving with large language models},
  author = {Yao, Shunyu and Yu, Dian and Zhao, Jeffrey and Shafran, Izhak and Griffiths, Tom and Cao, Yuan and Narasimhan, Karthik},
  journal = {Advances in neural information processing systems},
  volume = {36},
  pages = {11809--11822},
  year = {2023}
}

@inproceedings{gao2023pal,
  title = {Pal: Program-aided language models},
  author = {Gao, Luyu and Madaan, Aman and Zhou, Shuyan and Alon, Uri and Liu, Pengfei and Yang, Yiming and Callan, Jamie and Neubig, Graham},
  booktitle = {International conference on machine learning},
  pages = {10764--10799},
  year = {2023},
  organization = {PMLR}
}

@article{chen2022program,
  title = {Program of thoughts prompting: Disentangling computation from reasoning for numerical reasoning tasks},
  author = {Chen, Wenhu and Ma, Xueguang and Wang, Xinyi and Cohen, William W},
  journal = {arXiv preprint arXiv:2211.12588},
  year = {2022}
}

@article{madaan2023self,
  title = {Self-refine: Iterative refinement with self-feedback},
  author = {Madaan, Aman and Tandon, Niket and Gupta, Prakhar and Hallinan, Skyler and Gao, Luyu and Wiegreffe, Sarah and Alon, Uri and Dziri, Nouha and Prabhumoye, Shrimai and Yang, Yiming and others},
  journal = {Advances in neural information processing systems},
  volume = {36},
  pages = {46534--46594},
  year = {2023}
}

@article{shinn2023reflexion,
  title = {Reflexion: Language agents with verbal reinforcement learning},
  author = {Shinn, Noah and Cassano, Federico and Gopinath, Ashwin and Narasimhan, Karthik and Yao, Shunyu},
  journal = {Advances in neural information processing systems},
  volume = {36},
  pages = {8634--8652},
  year = {2023}
}

@article{brown2024large,
  title = {Large language monkeys: Scaling inference compute with repeated sampling},
  author = {Brown, Bradley and Juravsky, Jordan and Ehrlich, Ryan and Clark, Ronald and Le, Quoc V and R{\'e}, Christopher and Mirhoseini, Azalia},
  journal = {arXiv preprint arXiv:2407.21787},
  year = {2024}
}

@article{li2022competition,
  title = {Competition-level code generation with alphacode},
  author = {Li, Yujia and Choi, David and Chung, Junyoung and Kushman, Nate and Schrittwieser, Julian and Leblond, R{\'e}mi and Eccles, Tom and Keeling, James and Gimeno, Felix and Dal Lago, Agustin and others},
  journal = {Science},
  volume = {378},
  number = {6624},
  pages = {1092--1097},
  year = {2022},
  publisher = {American Association for the Advancement of Science}
}

@article{uesato2022solving,
  title = {Solving math word problems with process-and outcome-based feedback},
  author = {Uesato, Jonathan and Kushman, Nate and Kumar, Ramana and Song, Francis and Siegel, Noah and Wang, Lisa and Creswell, Antonia and Irving, Geoffrey and Higgins, Irina},
  journal = {arXiv preprint arXiv:2211.14275},
  year = {2022}
}

@inproceedings{wang2024math,
  title = {Math-shepherd: Verify and reinforce llms step-by-step without human annotations},
  author = {Wang, Peiyi and Li, Lei and Shao, Zhihong and Xu, Runxin and Dai, Damai and Li, Yifei and Chen, Deli and Wu, Yu and Sui, Zhifang},
  booktitle = {Proceedings of the 62nd Annual Meeting of the Association for Computational Linguistics (Volume 1: Long Papers)},
  pages = {9426--9439},
  year = {2024}
}

@inproceedings{lambert2025rewardbench,
  title = {Rewardbench: Evaluating reward models for language modeling},
  author = {Lambert, Nathan and Pyatkin, Valentina and Morrison, Jacob and Miranda, LJ and Lin, Bill Yuchen and Chandu, Khyathi and Dziri, Nouha and Kumar, Sachin and Zick, Tom and Choi, Yejin and others},
  booktitle = {Findings of the Association for Computational Linguistics: NAACL 2025},
  pages = {1755--1797},
  year = {2025}
}

@inproceedings{zheng2025processbench,
  title = {Processbench: Identifying process errors in mathematical reasoning},
  author = {Zheng, Chujie and Zhang, Zhenru and Zhang, Beichen and Lin, Runji and Lu, Keming and Yu, Bowen and Liu, Dayiheng and Zhou, Jingren and Lin, Junyang},
  booktitle = {Proceedings of the 63rd Annual Meeting of the Association for Computational Linguistics (Volume 1: Long Papers)},
  pages = {1009--1024},
  year = {2025}
}

@article{graves2016adaptive,
  title = {Adaptive computation time for recurrent neural networks},
  author = {Graves, Alex},
  journal = {arXiv preprint arXiv:1603.08983},
  year = {2016}
}

@article{chen2023frugalgpt,
  title = {Frugalgpt: How to use large language models while reducing cost and improving performance},
  author = {Chen, Lingjiao and Zaharia, Matei and Zou, James},
  journal = {arXiv preprint arXiv:2305.05176},
  year = {2023}
}

@inproceedings{jiang2023llm,
  title = {Llm-blender: Ensembling large language models with pairwise ranking and generative fusion},
  author = {Jiang, Dongfu and Ren, Xiang and Lin, Bill Yuchen},
  booktitle = {Proceedings of the 61st Annual Meeting of the Association for Computational Linguistics (Volume 1: Long Papers)},
  pages = {14165--14178},
  year = {2023}
}

@article{ong2024routellm,
  title = {Routellm: Learning to route llms with preference data},
  author = {Ong, Isaac and Almahairi, Amjad and Wu, Vincent and Chiang, Wei-Lin and Wu, Tianhao and Gonzalez, Joseph E and Kadous, M Waleed and Stoica, Ion},
  journal = {arXiv preprint arXiv:2406.18665},
  year = {2024}
}

@article{lewkowycz2022solving,
  title = {Solving quantitative reasoning problems with language models},
  author = {Lewkowycz, Aitor and Andreassen, Anders and Dohan, David and Dyer, Ethan and Michalewski, Henryk and Ramasesh, Vinay and Slone, Ambrose and Anil, Cem and Schlag, Imanol and Gutman-Solo, Theo and others},
  journal = {Advances in neural information processing systems},
  volume = {35},
  pages = {3843--3857},
  year = {2022}
}

@article{shao2024deepseekmath,
  title = {Deepseekmath: Pushing the limits of mathematical reasoning in open language models},
  author = {Shao, Zhihong and Wang, Peiyi and Zhu, Qihao and Xu, Runxin and Song, Junxiao and Bi, Xiao and Zhang, Haowei and Zhang, Mingchuan and Li, YK and Wu, Yang and others},
  journal = {arXiv preprint arXiv:2402.03300},
  year = {2024}
}

@article{rein2023gpqa,
  title = {Gpqa: A graduate-level google-proof q\&a benchmark},
  author = {Rein, David and Hou, Betty Li and Stickland, Asa Cooper and Petty, Jackson and Pang, Richard Yuanzhe and Dirani, Julien and Michael, Julian and Bowman, Samuel R},
  journal = {arXiv preprint arXiv:2311.12022},
  year = {2023}
}

@inproceedings{he2024olympiadbench,
  title = {Olympiadbench: A challenging benchmark for promoting agi with olympiad-level bilingual multimodal scientific problems},
  author = {He, Chaoqun and Luo, Renjie and Bai, Yuzhuo and Hu, Shengding and Thai, Zhen and Shen, Junhao and Hu, Jinyi and Han, Xu and Huang, Yujie and Zhang, Yuxiang and others},
  booktitle = {Proceedings of the 62nd Annual Meeting of the Association for Computational Linguistics (Volume 1: Long Papers)},
  pages = {3828--3850},
  year = {2024}
}

@inproceedings{jain2025livecodebench,
  title = {Livecodebench: Holistic and contamination free evaluation of large language models for code},
  author = {Jain, Naman and Gu, Alex and Li, Wen-Ding and Yan, Fanjia and Zhang, Tianjun and Wang, Sida and Solar-Lezama, Armando and Sen, Koushik and Stoica, Ion},
  booktitle = {International Conference on Learning Representations},
  volume = {2025},
  pages = {58791--58831},
  year = {2025}
}

@article{guo2025deepseek,
  title = {Deepseek-r1: Incentivizing reasoning capability in llms via reinforcement learning},
  author = {Guo, Daya and Yang, Dejian and Zhang, Haowei and Song, Junxiao and Wang, Peiyi and Zhu, Qihao and Xu, Runxin and Zhang, Ruoyu and Ma, Shirong and Bi, Xiao and others},
  journal = {arXiv preprint arXiv:2501.12948},
  year = {2025}
}

@article{zhang2025and,
  title = {What, how, where, and how well? a survey on test-time scaling in large language models},
  author = {Zhang, Qiyuan and Lyu, Fuyuan and Sun, Zexu and Wang, Lei and Zhang, Weixu and Guo, Zhihan and Wang, Yufei and King, Irwin and Liu, Xue and Ma, Chen},
  journal = {arXiv preprint arXiv:2503.24235},
  year = {2025}
}

@article{fu2024efficiently,
  title = {Efficiently serving llm reasoning programs with certaindex},
  author = {Fu, Yichao and Chen, Junda and Zhu, Siqi and Fu, Zheyu and Dai, Zhongdongming and Qiao, Aurick and Zhang, Hao},
  journal = {arXiv e-prints},
  pages = {arXiv--2412},
  year = {2024}
}

@article{huang2025latency,
  title = {Latency and Token-Aware Test-Time Compute},
  author = {Huang, Jenny Y and Damani, Mehul and El-Kurdi, Yousef and Astudillo, Ramon and Sun, Wei},
  journal = {arXiv preprint arXiv:2509.09864},
  year = {2025}
}

\section*{Supplementary Material}
\appendix

\section{Appendix A: Detailed Experimental Protocol}

\subsection{Candidate Generation}

For each dataset--generator pair, we generated a shared candidate pool with $N=16$ candidates per example and a maximum generation budget of 16,384 output tokens.  The final replay uses the 16,384-token pool because it contains the widest candidate support for both generation-only and verifier-routed policies.  Earlier 4,096-token pools were used during development and quality checks but are not mixed into the main table.

The generators were Qwen3-14B, Phi-4-reasoning, and Qwen3-8B.  Sampling used temperature 0.6 and top-$p=0.95$ for multi-sample candidates; greedy and fixed-long baselines use the first deterministic candidate in the replay pool.  All generated text, extracted answers, token counts, and cap indicators were retained for audit.

\subsection{Verifier Tiers}

The verifier tiers are:
\begin{itemize}
  \item $V_0$: rule-based answer frequency and normalized answer agreement.
  \item $V_1$: Qwen3-8B local outcome judge, applied to all final candidates used in replay.
  \item $V_2$: Phi-4-reasoning step evaluator, scored once for warm-up candidates and not for candidates added during the adaptive sampling loop, because process parsing was sparse in the final run; treated as auxiliary.
  \item $V_3$: Qwen3-14B deep outcome verifier, routed to top candidates according to $V_1$ and answer-frequency ranking.
\end{itemize}

\subsection{Answer Extraction and Quality Control}

Mathematical answer extraction prioritizes boxed answers, explicit final-answer statements, and final symbolic expressions.  The enhanced extractor normalizes fractions, tuples, common LaTeX forms, plain-text $\pi$, and comma-separated answer sets.  We add audit fields for re-extracted answer, extraction status, token-cap hits, repetition score, and incomplete boxed expressions.  These fields are used for evaluation and analysis while preserving raw model outputs.

\section{Appendix B: Additional Results}
\label{app:additional-results}

\begin{figure}[!t]
  \centering
  \includegraphics[width=0.88\columnwidth]{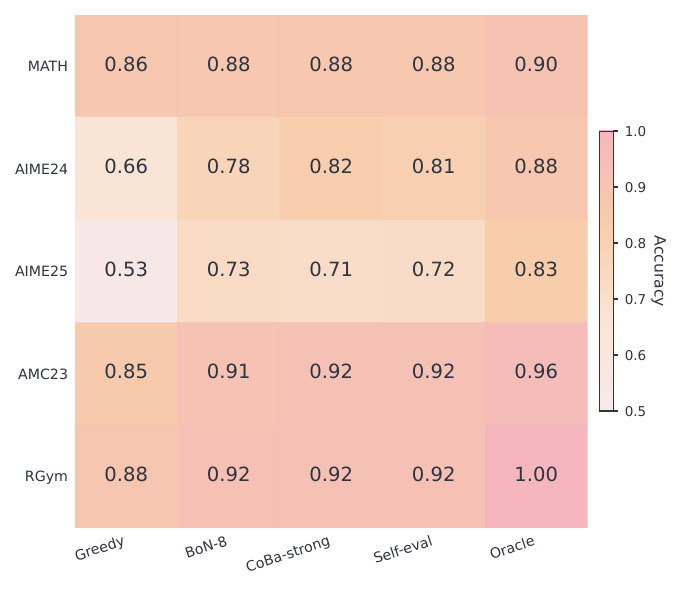}
  \caption{Per-dataset accuracy heatmap.  Values are averaged over the three generators; the oracle is an upper bound.}
  \label{fig:heatmap}
\end{figure}

\begin{figure}[!t]
  \centering
  \includegraphics[width=0.92\columnwidth]{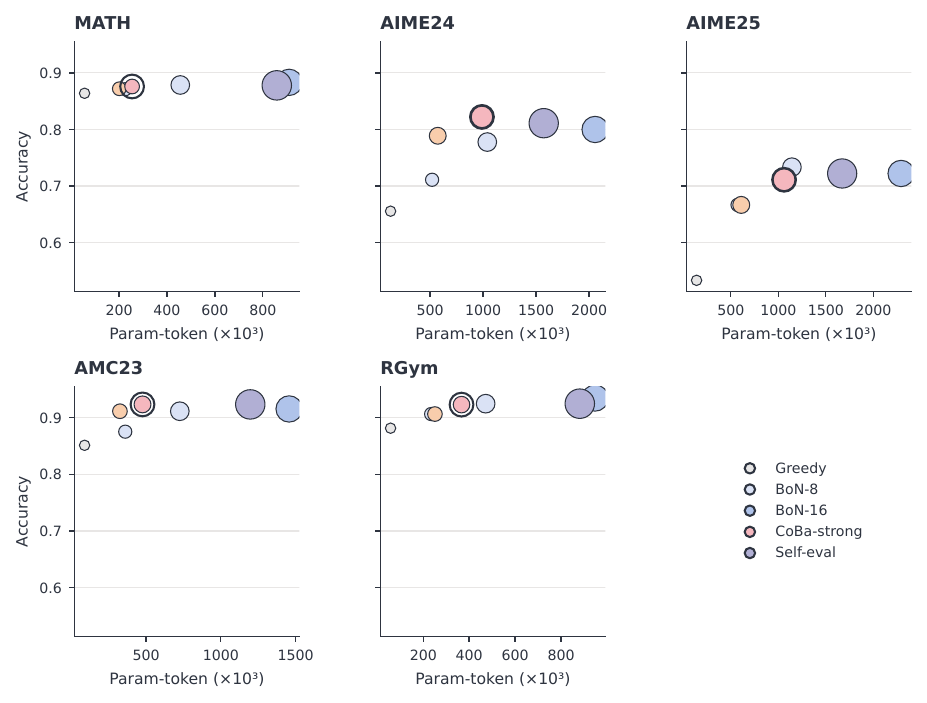}
  \caption{Dataset-level Pareto frontiers.  Routed policies improve the frontier across the main task groups, with the largest remaining gap on AIME 2025.}
  \label{fig:smallmultiples}
\end{figure}

\begin{figure}[!t]
  \centering
  \includegraphics[width=0.93\columnwidth]{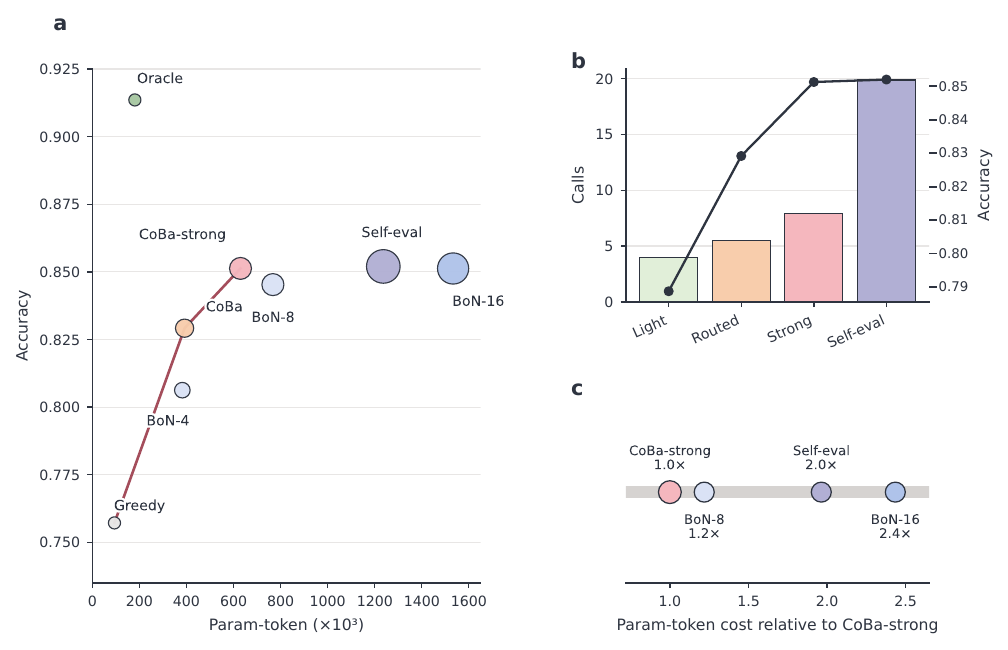}
  \caption{Evidence grid for compute allocation.  The panels show frontier movement, routed verification strength, and relative cost against high-accuracy baselines.}
  \label{fig:evidence_supp}
\end{figure}

\begin{figure}[!t]
  \centering
  \includegraphics[width=0.93\columnwidth]{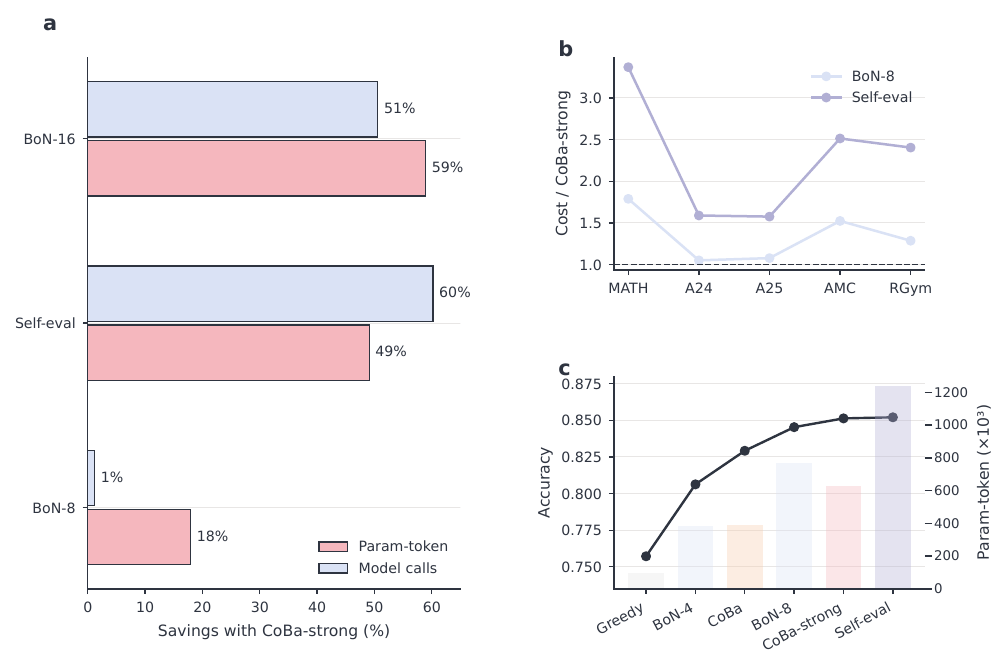}
  \caption{Cost savings at comparable accuracy.  \strong{} obtains the accuracy of strong sampling or self-evaluation baselines with fewer total tokens and parameter-weighted tokens.}
  \label{fig:savings_supp}
\end{figure}

Figure~\ref{fig:heatmap} visualizes the per-dataset accuracy table as a compact heatmap.  Figure~\ref{fig:evidence_supp} gives the evidence grid that was moved out of the main text to reduce cross-column fragmentation.  Figure~\ref{fig:smallmultiples} provides the dataset-level frontier breakdown that supports the main-text allocation analysis; the action-allocation audit now appears in the main paper.  Figure~\ref{fig:savings_supp} gives the compact cost-savings view that was moved out of the main text to keep the central narrative focused on the frontier, per-dataset evidence, and the main cost table.  The released experiment artifacts additionally contain the complete per-generator tables, complete pairwise significance matrix, and cost-analysis tables used to produce the figures.

\subsection{Additional Statistical Details}

The main text reports the paired bootstrap comparisons most relevant to the central claim.  Full pairwise tests are retained in the released experiment artifacts and include every replayed baseline pair.  We avoid placing the complete matrix in the appendix because it is large and mostly redundant with the focused comparisons summarized in the main text.

\section{Appendix C: Baseline Definitions}

\textbf{Greedy / fixed long.}
The first candidate in the replay pool is used as the single-sample answer.  In the final replay, fixed-long and greedy coincide because the stored pool is generated at the long budget and both select the first candidate without additional verification.

\textbf{Best-of-$N$ majority and self-consistency.}
For $N\in\{2,4,8,16\}$, the first $N$ candidates are grouped by normalized final answer.  The most frequent answer is selected; ties are broken by the earliest candidate.  Self-consistency is identical in the final replay because all candidates contain reasoning traces.

\textbf{Evaluator-scaling proxies.}
Outcome and process-plus-outcome evaluator baselines score candidate prefixes with local judges and select the highest-scoring answer.  They are labeled as proxies because they use local Qwen3/Phi-style evaluators rather than a separately trained reward model checkpoint.

\textbf{Difficulty-adaptive and uncertainty allocation proxies.}
These baselines allocate more candidate slots to examples with higher answer disagreement or lower judge confidence.  They preserve the core adaptive-compute intuition but are not claimed as exact reproductions of any unpublished or checkpoint-dependent method.

\textbf{Self-evaluation weighted voting.}
The generator's candidate answers are weighted by local outcome-judge scores and answer frequency.  This baseline approximates the inference-time pattern of unified reasoner-verifier systems, but we do not claim to reproduce any full RL-trained verifier.

\textbf{Pool oracle.}
The pool oracle selects a correct candidate if any candidate in the offline pool is correct.  It is included only to quantify headroom and is never compared as a deployable method.

\section{Appendix D: Reproducibility Notes}

All main methods are replayed from the same generated candidate pools.  This design makes the comparison conservative: routing can change which candidates are inspected and selected, but it cannot benefit from private generations unavailable to the baselines.  Verifier outputs are likewise cached and shared across methods when the same verifier tier is used.  The pool oracle is the only method allowed to inspect correctness labels, and it is reported only as a headroom estimate.

The final tables report macro averages over dataset--generator pairs.  Per-generator and per-dataset records are retained in the experiment artifacts, together with the answer-extraction audit fields used during quality control.  The released logs distinguish raw extracted answers from enhanced extracted answers so that extraction changes can be audited without modifying the original model completions.

Several baselines are necessarily local proxies for broader method families.  The self-evaluation weighted-voting baseline captures the inference pattern of using the same local model family for reasoning and verification, but it is not claimed as a reproduction of a separately RL-trained verifier.  Similarly, the evaluator-scaling baselines use local Qwen3 and Phi-style evaluators rather than closed or unavailable reward-model checkpoints.  These choices keep the comparison reproducible on the same local hardware stack as \method{}.
\end{document}